\documentclass[10pt,twocolumn,letterpaper]{article}

\usepackage{cvpr}              

\usepackage{tabularx}
\usepackage{booktabs}
\usepackage{makecell}
\usepackage{booktabs}
\usepackage{colortbl}
\usepackage{xcolor}
\usepackage{graphicx}
\usepackage{subcaption}
\usepackage{dblfloatfix}
\usepackage{graphicx}
\usepackage{caption}
\usepackage{subcaption}

\definecolor{cvprblue}{rgb}{0.21,0.49,0.74}
\usepackage[pagebackref,breaklinks,colorlinks,allcolors=cvprblue]{hyperref}

\def\paperID{27} 
\def\confName{CVPR}
\def\confYear{2026}

\title{Understanding Identity Continuity in Thermal Video through Scene-Level Consistency}

\author{Wei-Chieh Sun, Gyungmin Ko, Heejae Kwon, Hsiang-Wei Huang, Jenq-Neng Hwang \\
Department of Electrical and Computer Engineering,\\
Information Processing Lab, University of Washington, USA\\
{\tt\small \{wsun12, gmko, hkwon4, hshuang, hwang\}@uw.edu}
}

\begin{document}

\maketitle

\begin{abstract}
Thermal pedestrian MOT remains challenging because weak appearance cues and frequent detection interruptions cause severe trajectory fragmentation. We study whether lightweight post-processing can recover identity continuity without relying on heavy re-identification models or complex online association. Starting from a YOLOv8 and SORT baseline, we add a modular identity-repair backend consisting of online short-gap remapping and offline tracklet relinking based on temporal, spatial, motion, and border cues. Controlled ablations on a fixed validation split and evaluation on the official PBVS Thermal Pedestrian MOT benchmark show that the main identity gains arise from conservative relinking, improving IDF1 from 82.25 to 84.93 while preserving MOTA, whereas many heuristic thresholds remain stable across broad operating ranges. These results suggest that, in low-information thermal imagery, robust identity recovery can be achieved more effectively through high-precision trajectory relinking than through increasing tracker complexity. These results provide a controlled analysis of identity recovery in thermal video, showing that scene-level spatial-temporal consistency plays a dominant role in identity continuity compared to local frame-to-frame association.\footnote{Project page: \url{https://heejaeee.github.io/pbvs26_tmot/}}
\end{abstract}

\section{Introduction}
\label{sec:intro}

Multi-object tracking (MOT) in thermal imagery has become increasingly important for surveillance, public safety, intelligent transportation, and other vision systems that must operate reliably under poor illumination, adverse weather, or low-visibility conditions. Unlike RGB cameras, thermal sensors capture infrared radiation and therefore remain effective when color and texture cues are severely degraded. This property makes thermal imaging especially attractive for pedestrian monitoring in nighttime and safety-critical environments. At the same time, thermal pedestrian tracking remains difficult because objects often appear with weak boundaries, limited appearance variation, and highly ambiguous local structure, all of which make detection and identity preservation substantially more challenging than in conventional RGB settings.

Despite steady progress in modern MOT, thermal-based tracking continues to present several unresolved difficulties. First, the low-information nature of thermal imagery reduces the discriminative power of appearance features, making it harder to distinguish nearby pedestrians or recover identities after occlusion. Second, crowded scenes, missed detections, and abrupt motion changes frequently fragment trajectories and induce identity switches. Third, many high-performing tracking systems depend on heavy re-identification modules or complex motion models, which can improve robustness but often increase computational cost and deployment complexity. These limitations are particularly restrictive when the goal is to build a practical system that is both accurate and efficient for real-time thermal pedestrian tracking.

To study this problem, we develop a lightweight thermal MOT baseline centered on detector-driven tracking and explicit identity repair. The framework combines a thermal-adapted YOLOv8 detector with a SORT tracker and a modular post-processing backend for recovering fragmented identities. Our primary goal is to identify which mechanisms drive identity continuity in thermal scenes, not to introduce a new tracking architecture. In particular, we treat identity continuity as a scene-level phenomenon that emerges from consistent spatial-temporal structure across frames, rather than as a purely local data association problem. This reframing connects identity continuity in thermal tracking to a broader question in video understanding: how global scene structure constrains and stabilizes object identity over time under weak visual signals. In particular, we isolate identity continuity as a property of scene-level consistency instead of a byproduct of architectural complexity.

In brief, the main contributions are as follows:
\begin{itemize}
\item We provide a controlled analysis showing that identity continuity in thermal video is primarily governed by conservative spatial-temporal relinking rather than online tracker complexity.
\item We demonstrate through component ablation that long-range relinking, rather than short-gap stitching, is the dominant factor in recovering fragmented identities.
\item We perform a sensitivity study showing that identity recovery remains stable across broad threshold ranges, indicating that the effect is not dependent on narrow parameter tuning.
\item We introduce a reproducible experimental framework for studying identity recovery as a scene-level consistency problem in low-information video.
\end{itemize}

In addition to controlled ablations on a fixed validation split, we report results on the official PBVS benchmark evaluation server to ensure comparability with prior work.

\begin{figure*}
  \centering
  \includegraphics[width=\linewidth]{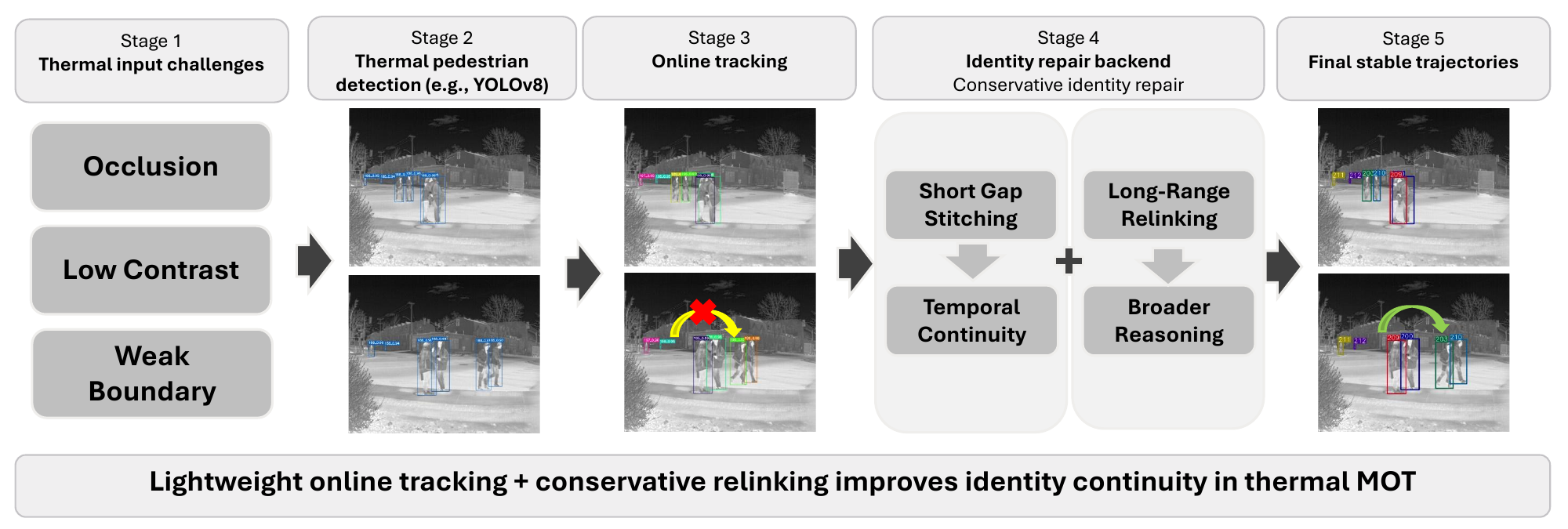}
  \vspace{0.5em}
  \caption{Overview of the proposed framework for identity continuity in thermal video. The pipeline begins with challenging thermal inputs characterized by occlusion, low contrast, and weak boundaries, followed by thermal pedestrian detection and lightweight online tracking. A conservative identity-repair backend then recovers fragmented trajectories through auxiliary short-gap stitching and dominant long-range relinking, yielding final stable identity-consistent trajectories.}
  \label{fig:pipeline}
\end{figure*}

\section{Related Work}
\label{sec:relatedwork}

Modern multi-object tracking (MOT) systems are largely organized around the \emph{tracking-by-detection} paradigm, where an object detector first produces per-frame observations and a tracker subsequently enforces temporal identity consistency. Over the past decade, a diverse family of association strategies, motion models, and auxiliary cues such as appearance or segmentation have been proposed to improve identity stability under occlusion and detection noise. This section reviews the most relevant strands of prior work and situates our system within this landscape.

\subsection{Tracking-by-Detection}

A foundational approach in modern MOT is SORT~\cite{bewley2016sort}, which combines Kalman filtering with Hungarian assignment based on geometric overlap. Despite its simplicity, SORT remains widely used due to its efficiency and modular design, making it a natural baseline for detector-centered tracking systems. Subsequent work has focused on strengthening data association while retaining the same basic structure. ByteTrack~\cite{zhang2022bytetrack} demonstrated that discarding low-confidence detections prematurely can significantly degrade identity continuity; instead, these detections can be recovered in a secondary matching stage to reduce track fragmentation under occlusion. BoT-SORT~\cite{aharon2022botsort} further enhances this paradigm by integrating improved motion modeling, global motion compensation, and optional appearance embeddings to improve robustness in complex scenes. In parallel, OC-SORT~\cite{cao2023ocsort} revisits the Kalman update formulation by emphasizing observation-centric updates, improving tracking stability under non-linear motion and short-term occlusions. Beyond classical tracking-by-detection, recent work has explored jointly learned detection and association mechanisms. For example, CenterTrack formulates tracking as point-based detection and temporal offset prediction~\cite{zhou2020tracking}. Such methods reduce reliance on hand-crafted association logic, but often assume visual cues and training regimes that differ from low-contrast thermal pedestrian settings. Our work is complementary: rather than proposing a new jointly trained tracker, we study how much identity continuity can be recovered through lightweight trajectory-level repair in thermal imagery. Accordingly, our goal is not to outperform fully learned end-to-end trackers in general settings, but to establish a strong and analyzable lightweight baseline for low-visibility thermal MOT.

\subsection{Identity Recovery and Tracklet Stitching}
To address identity switches, where a tracker assigns different IDs to the same target over time, prior work has proposed identity-recovery and tracklet-stitching methods that associate fragmented tracklets using appearance, spatial, and temporal cues. These methods typically rely on motion or appearance features extracted from tracklets for tracklet-level association. For example, Translink \cite{zhang2023translink} incorporates a CNN and a temporal attention network to extract and encode the appearance features of a tracklet and formulates the merging of tracklet pairs as a binary classification task. AFLink \cite{du2023strongsort} relies only on spatial and temporal information. Some methods \cite{huang2023enhancing, yang2024online, cherdchusakulchai2024online} adopt feature clustering techniques to merge tracklets and improve performance in multi-camera tracking scenarios. MambaMOT \cite{huang2025mambamot} proposes a motion model that acts as a motion predictor and extracts tracklet motion features for tracklet association. Recent work has also emphasized the importance of strong baselines and careful evaluation in thermal infrared tracking~\cite{Stadler_2025_CVPR}. Our work aligns with this perspective. Rather than introducing a heavy end-to-end architecture, we focus on a lightweight detector-centered baseline and analyze which repair cues meaningfully improve identity continuity under thermal fragmentation.

\section{Method}
\label{sec:method}

Our formulation isolates how scene-level spatial and temporal consistency can recover identity continuity from fragmented observations. The goal is not to maximize tracking performance through architectural complexity, but to identify which spatial-temporal cues are sufficient to explain identity recovery in low-information thermal video. In our validation configuration, the system processes sequences from the PBVS TP-MOT benchmark~\cite{elahmar2025tpmot} at 10 FPS with resolution $960\times1280$. The overall design is intentionally modular: detections are first generated and cached, and trajectories are then estimated online. Identity continuity is subsequently refined by an offline repair backend. Within this backend, short-gap stitching serves as an auxiliary recovery mechanism, while conservative longer-range relinking provides the dominant empirical gain. This decomposition follows the tracking-by-detection philosophy of SORT~\cite{bewley2016sort} while adapting it to the fragmentation patterns that are common in thermal imagery~\cite{tran2025noveltuningmethodrealtime}.

\paragraph{Thermal pedestrian detection.}
For each frame $I_t$, we apply a YOLOv8 detector~\cite{jocher2024ultralytics, wang2026detectorintheloop} in a single-class setting (\emph{person}). The detector operates on resized inputs of $1920\times1920$, uses batched inference with batch size 10, and is configured with a very low confidence threshold ($10^{-4}$) and non-maximum suppression threshold 0.75 in order to preserve recall. Let
\[
\mathcal{D}_t=\{(b_i^t,s_i^t)\}_{i=1}^{N_t}
\]
denote the set of normalized bounding boxes and confidence scores predicted at time $t$. Instead of directly coupling detector and tracker in memory, we store $\mathcal{D}_t$ as per-frame YOLO text files. This choice improves reproducibility, allows detector/tracker ablations without re-running the full pipeline, and enables direct inspection of detector failure modes.

\paragraph{Online tracking.}
Before tracking, the stored detections are converted from normalized YOLO coordinates to image-space $(x_1,y_1,x_2,y_2)$ boxes and filtered with a stricter threshold $s_i^t \geq 0.5$. Online association is then performed with SORT~\cite{bewley2016sort}, which combines a constant-velocity Kalman predictor with Hungarian matching over bounding-box overlap. Given predicted track boxes $\hat{\mathcal{T}}_t$ and detections $\mathcal{D}_t$, the association stage maximizes pairwise IoU, or equivalently minimizes
\[
C_{ij}=1-\mathrm{IoU}(\hat{b}_j^t,b_i^t).
\]
In our experimental setting, tracks are retained for up to 40 missed frames, and the IoU gate is set to 0.001, which indicates a deliberately permissive association policy. This setting is useful in thermal scenes where shape deformation, low contrast, and partial occlusion can make frame-to-frame overlap unstable even when identity continuity is still visually plausible.

\paragraph{Online short-gap identity repair.}
A key implementation detail is that identity correction is not deferred entirely to the end of the sequence. Immediately after each SORT update, we apply an online remapping module that keeps a memory of recently lost output identities. If a newborn raw track $u$ appears within $\Delta t \leq 7$ frames of a recently lost track $v$, and if their centers satisfy
\[
\|c_u-c_v\|_2 \leq 60,
\]
the newborn trajectory inherits the previous output identity rather than emitting a new ID. Because the benchmark is single-class, the optional label-consistency check is effectively always satisfied. This step is particularly effective for thermal videos, where brief detector interruptions often arise from local temperature ambiguity, partial truncation, or low signal-to-noise regions. These values correspond to short-term interruptions at 10 FPS and are chosen to capture plausible brief occlusions without allowing aggressive identity propagation. As confirmed by the component ablation in Sec~\ref{sec:ablation}, this stage improves short-term continuity in the raw tracker, while the dominant final gain remains attributable to the offline relinking stage.

\paragraph{Offline motion-consistent stitching.}
The first offline post-processing stage explicitly targets short illegal disappear/reappear events. We treat a tracklet ending away from the image border as an implausible disappearance and a tracklet starting away from the border as an implausible appearance. For an old tracklet $a$ and a new tracklet $b$, with temporal gap
\[
\Delta t=t_b^{\mathrm{start}}-t_a^{\mathrm{end}},
\]
we estimate endpoint velocity from a 3-point temporal window and extrapolate the previous trajectory as
\[
\tilde{c}_a=c_a^{\mathrm{end}}+v_a\Delta t.
\]
We merge $b$ into $a$ only when all of the following hold: 1) both endpoints lie outside a 60-pixel border band, 2) $1\leq \Delta t \leq 30$, 3) the predicted displacement satisfies $\|\tilde{c}_a-c_b^{\mathrm{start}}\|_2 \leq 80$ pixels, and 4) if the motion magnitude is informative (speed $\geq 0.25$ pixels/frame), the angle between the two velocity vectors is at most $45^\circ$ and the speed ratio is at most 3.0. This stage is intentionally conservative: it only repairs fragments that are both temporally short and kinematically consistent. These thresholds encode conservative kinematic plausibility, limiting merges to short gaps and physically reasonable displacements.

\paragraph{Conservative relinking of longer fragments.}
A second offline stage addresses longer-range fragmentation using a fixed logistic relinking function. For each candidate predecessor--successor pair, we construct a 48-dimensional feature vector that encodes temporal gap, Euclidean and axis-wise displacement, tracklet lengths, box-size ratios, border distances, edge indicators, global velocity consistency, and multi-window extrapolation errors computed over temporal windows $\{2,3,5,10,20\}$. A candidate pair is considered only if the gap is at most 60 frames, the endpoint distance is at most 120 pixels, and both tracklets contain at least two observations; moreover, the predecessor must not terminate near a 25-pixel image border. The final relinking probability is
\[
p=\sigma(w^\top z),
\]
where \(z\) is the normalized feature vector and \(\sigma(\cdot)\) is the sigmoid function. In our implementation, the weight vector \(w\) is fixed throughout evaluation and used only as a lightweight scoring function over hand-crafted tracklet features. We accept a merge only when $p \geq 0.95$ and the candidate is mutually optimal for both the old and new tracklet. This mutual-best constraint makes the relinking pass high-precision and prevents cascade errors that would otherwise amplify early mistakes. The relinking thresholds are intentionally conservative and are further examined through sensitivity analysis in Sec.~\ref{sec:ablation}.

\paragraph{Output formatting and reproducibility.}
After post-processing, each visible target is written in MOT-style format as
\[
(f,\mathrm{id},x,y,w,h,s,\ell,-1,-1),
\]
where $f$ is the 1-indexed frame number, $\mathrm{id}$ is the repaired identity, $(x,y,w,h)$ is the image-space bounding box, $s$ is the confidence score, and $\ell$ is the class label. The implementation optionally renders qualitative overlays during detection, tracking, and post-processing for evaluation. 

Overall, the system can be interpreted as a high-recall thermal detector, a lightweight online motion tracker, and a two-stage identity-repair backend. This asymmetry is well suited to thermal MOT: the detector maximizes candidate coverage, the online tracker preserves short-term temporal continuity, and the post-processing stages recover long-range identity consistency under missed detections, brief occlusions, and mid-frame fragmentation. Because the repair backend contains several interpretable geometric thresholds, we explicitly evaluate both component-wise contribution and threshold sensitivity in Sec.~\ref{sec:ablation}.

\begin{table}[t]
\centering
\caption{PBVS TP-MOT split statistics used in this paper (computed from local annotation files).}
\label{tab:split_stats}
\small
\resizebox{\columnwidth}{!}{
\begin{tabular}{lcccc}
\toprule
Split & \#Seq & \#Frames & \#Boxes & Density (mean / P95 / max ped/frame) \\
\midrule
Train & 24 & 7,150 & 38,111 & 5.33 / 9 / 14 \\
Val  & 6 & 1,683 & 13,552 & 8.05 / 17 / 20 \\
\bottomrule
\end{tabular}
}
\end{table}

\section{Experiments}
\subsection{Evaluation Protocol}
\label{sec:eval_protocol}
We evaluate the proposed analysis on the PBVS Thermal Pedestrian MOT (TP-MOT) benchmark~\cite{elahmar2025tpmot}, the first large-scale thermal imaging dataset annotated specifically for pedestrian multi-object tracking. The dataset was captured using a FLIR ADK thermal sensor at five urban intersections, comprising 30 sequences with a total of 9{,}000 frames at a native resolution of $960\times1280$ and a frame rate of 10 FPS. The sequences contain single-class pedestrian annotations and exhibit common thermal tracking challenges including occlusion, low contrast, and trajectory fragmentation. For our controlled ablation experiments, we use a fixed local validation split consisting of six sequences (Seq2, Seq17, Seq22, Seq47, Seq54, Seq66), selected to cover diverse scene layouts and pedestrian densities. This split is held consistent across all tracker variants and repair configurations to isolate the effect of identity-repair components. In addition to the local validation split used for controlled ablation, we evaluate the proposed method on the official PBVS TP-MOT benchmark evaluation protocol. Results are reported using the standard benchmark evaluation server to ensure comparability with prior work. All claims regarding overall tracking performance are based on the official benchmark evaluation, while the local split is used exclusively for controlled analysis of identity-repair components. The dataset includes diverse urban conditions with varying pedestrian density, occlusion levels, and thermal contrast.

\subsection{Evaluation Metrics}
\label{sec:metrics}

We evaluate tracking performance using standard CLEAR MOT and identity-aware metrics that are widely used in multi-object tracking benchmarks and implemented in MOTChallenge-style evaluation toolchains \cite{bernardin2008evaluating,milan2016mot16}. In particular, we report MOTA, MOTP, IDF1, IDP, IDR, Recall, and Precision. These metrics jointly measure localization quality and temporal identity consistency.

Let $TP$, $FP$, and $FN$ denote the numbers of true positives, false positives, and false negatives aggregated over all frames. Let $IDSW$ denote the number of identity switches, and $GT$ denote the total number of ground-truth objects across all frames. For identity-aware evaluation, $IDTP$, $IDFP$, and $IDFN$ denote identity true positives, identity false positives, and identity false negatives, respectively.

\paragraph{MOTA.}
Multi-Object Tracking Accuracy summarizes the dominant tracking errors into a single score \cite{bernardin2008evaluating}:
\begin{equation}
\mathrm{MOTA} = 1 - \frac{FN + FP + IDSW}{GT}.
\end{equation}
A higher MOTA indicates fewer missed detections, fewer false alarms, and fewer identity switches. Because it penalizes all three error sources simultaneously, MOTA is commonly used as an overall measure of tracking performance.

\paragraph{MOTP.}
Multi-Object Tracking Precision measures localization quality on matched object--hypothesis pairs \cite{bernardin2008evaluating}. In overlap-based evaluation, it reflects the spatial agreement between matched predictions and ground-truth objects across all true-positive associations. In many benchmarks, including PBVS, MOTP is reported in a lower-is-better localization-error form.

\paragraph{IDP, IDR, and IDF1.}
Identity metrics evaluate whether detections are assigned the correct trajectory identity over time \cite{ristani2016performance}. Using identity counts,
\begin{equation}
\mathrm{IDP} = \frac{IDTP}{IDTP + IDFP},
\end{equation}

\begin{equation}
\mathrm{IDR} = \frac{IDTP}{IDTP + IDFN},
\end{equation}

\begin{equation}
\mathrm{IDF1} =
\frac{2\,IDTP}{2\,IDTP + IDFP + IDFN}
\end{equation}

IDP measures identity precision, IDR measures identity recall, and IDF1 is their harmonic mean. These metrics are particularly sensitive to trajectory fragmentation and identity switches, making them informative for evaluating identity preservation over time.

\paragraph{Recall and Precision.}
Detection Recall and Precision are defined as

\begin{equation}
\mathrm{Recall} = \frac{TP}{TP + FN},
\end{equation}

\begin{equation}
\mathrm{Precision} = \frac{TP}{TP + FP}.
\end{equation}

Recall measures the proportion of ground-truth objects that are successfully detected, while Precision measures the proportion of predicted detections that correspond to true objects. These metrics describe detection coverage and false-alarm behavior within the tracking pipeline \cite{milan2016mot16}.

\begin{figure*}
  \centering
  \includegraphics[width=\linewidth]{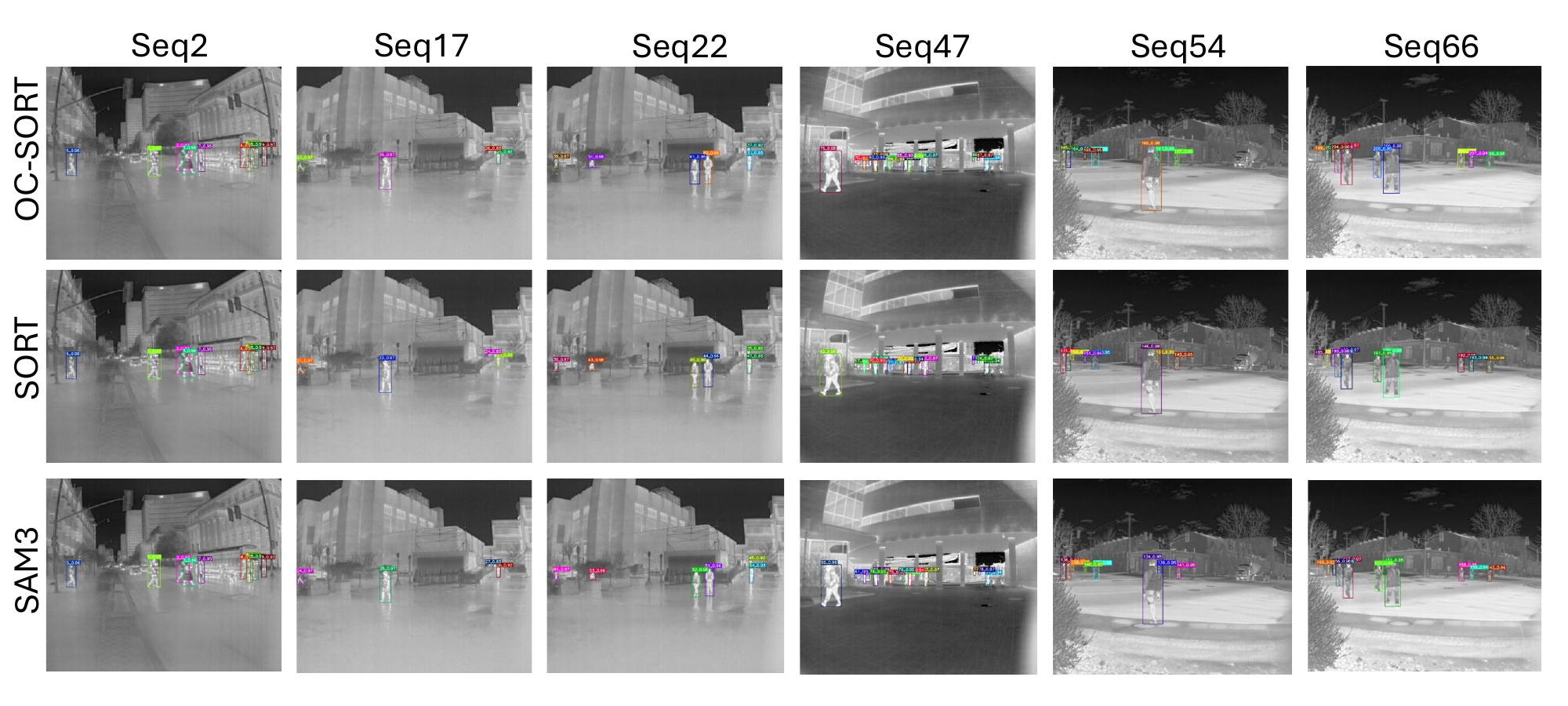}
  \vspace{0.5em}
  \caption{Qualitative comparison of OC-SORT, SORT, SAM3, and the proposed identity-repair method on PBVS TP-MOT sequences. The examples illustrate that enforcing scene-level spatial-temporal consistency through identity repair yields fewer spurious identity breaks and smoother trajectories under occlusion and motion changes.}
  \label{fig:qualitative}
\end{figure*}

\subsection{Experimental Setup}

Experiments are conducted on the PBVS TP-MOT benchmark described in Sec.~\ref{sec:eval_protocol}. All experiments follow a consistent evaluation protocol across tracker variants. The detector is a YOLOv8 model trained for single-class pedestrian detection, and we adopt the released weights from~\cite{tran2025noveltuningmethodrealtime}. Inference is performed on resized $1920\times1920$ frames. Detection results are stored and reused across tracker variants to ensure fair comparison. The tracking stage is evaluated using several widely used MOT association strategies implemented within the same pipeline, including SORT~\cite{bewley2016sort}, ByteTrack~\cite{zhang2022bytetrack}, BoT-SORT~\cite{aharon2022botsort}, OC-SORT~\cite{cao2023ocsort}, BoostTrack~\cite{stanojevic2024boosttrack}, DiffMOT~\cite{lv2024diffmot}, and a segmentation-assisted variant based on SAM3~\cite{carion2026sam}. We emphasize that the proposed identity-repair backend is applied only to the SORT baseline in this work. Extending the same post-processing to other trackers (e.g., ByteTrack, BoT-SORT) is feasible but is beyond the scope of this study. Therefore, the comparisons in Table~\ref{tab:official_baseline} should be interpreted as contextual rather than strictly controlled comparisons.

We report two types of results. Table~\ref{tab:official_baseline} provides contextual comparison with representative trackers under a common detection setting. In contrast, all component ablations and threshold-sensitivity studies in this paper are conducted on the local validation split using fixed detections and a unified evaluation script. Unless otherwise noted, our conclusions about component importance and threshold robustness are drawn from the controlled local validation experiments.

\begin{table}[t]
\centering
\caption{Contextual comparison with representative tracking methods under a common detection setting. 
All methods are evaluated without the proposed identity-repair backend. 
Our post-processing is applied only to the SORT baseline for controlled analysis and is not applied to other trackers in this table.}
\label{tab:official_baseline}
\small
\begin{tabular}{lccc}
\toprule
Method & MOTA $\uparrow$ & MOTP $\downarrow$ & IDF1 $\uparrow$ \\ 
\midrule
ByteTrack & 91.73 & 13.67 & 76.59 \\
BoT-SORT  & 91.74 & 13.68 & 76.05 \\
BoostTrack& 86.54 & 15.55 & 75.45 \\
DiffMOT   & 86.30 & 16.11 & 78.12 \\
OC-SORT   & 90.71 & 12.36 & 56.85 \\
SAM3      & 90.03 & 21.04 & 60.73 \\
SORT      & \textbf{98.44} & \textbf{12.63} & \textbf{82.25} \\
\bottomrule
\end{tabular}
\end{table}

\subsection{Comparison with Existing Trackers}

Table~\ref{tab:official_baseline} provides contextual comparison with representative tracking methods. Within this setting, SORT serves as a strong lightweight baseline, and adding identity repair improves identity continuity. While some trackers achieve strong detection recovery, their identity consistency under this evaluation setting appears lower than the SORT baseline. However, we emphasize that these methods are evaluated without the proposed post-processing backend, and therefore this comparison is not strictly controlled. These observations motivate the controlled ablations that follow, where we isolate which repair components are responsible for the gain.

\subsection{Component Ablation of Identity Repair}
\label{sec:ablation}

To identify which parts of the repair backend are responsible for the observed gains, we perform a controlled component ablation on the local validation split. Table~\ref{tab:component_ablation} compares the raw SORT baseline, isolated repair modules, and cue-removal variants under a fixed detector and evaluation protocol. This is important because it shows that the proposed gain does not arise from a loosely defined multi-stage pipeline, but from a specific and interpretable repair mechanism whose contribution can be isolated empirically.

\begin{table*}[t]
\centering
\caption{Component ablation of the proposed identity-repair backend on the local validation split. The main gain comes from the relinking stage, while stitch-only produces no measurable improvement on this split. Removing spatial consistency causes the largest degradation, indicating that spatial plausibility is the dominant cue for recovering fragmented thermal trajectories.}
\label{tab:component_ablation}
\small
\resizebox{\textwidth}{!}{
\begin{tabular}{lcccc}
\toprule
Variant & MOTA $\uparrow$ & IDF1 $\uparrow$ & MOTP-IoU $\uparrow$ & $\Delta$IDF1 vs. raw (pp) \\
\midrule
Raw SORT baseline & 98.44 & 82.25 & 87.37 & 0.00 \\
Stitch only & 98.46 & 82.25 & 87.37 & 0.00 \\
Relink only & 98.54 & 84.93 & 87.37 & +2.68 \\
Full pipeline (stitch + relink) & 98.54 & 84.93 & 87.37 & +2.68 \\
\midrule
Full w/o motion cue & 98.55 & 83.86 & 87.37 & +1.60 \\
Full w/o border cue & 98.54 & 84.54 & 87.37 & +2.29 \\
Full w/o temporal cue & 98.53 & 84.13 & 87.37 & +1.88 \\
Full w/o spatial cue & 98.52 & 80.50 & 87.37 & -1.76 \\
\midrule
Online ID repair off (raw) & 98.40 & 80.98 & 87.37 & -1.27 \\
Online ID repair off + offline full pipeline & 98.55 & 84.94 & 87.37 & +2.69 \\
\bottomrule
\end{tabular}
}
\end{table*}

The results show that the dominant improvement comes from the relinking stage. Relative to the raw SORT baseline, relinking improves IDF1 from 82.25 to 84.93 while preserving MOTA. In contrast, stitch-only produces no measurable gain on this validation split, indicating that the observed improvement is driven primarily by conservative long-range relinking rather than by short-gap stitching alone. This suggests that, for the studied sequences, the dominant fragmentation patterns are better characterized as longer-range association failures than as short-gap kinematic discontinuities. More importantly, it indicates that identity recovery in thermal video is driven primarily by longer-range scene-consistent reasoning rather than by local temporal continuity alone.

We also evaluate the contribution of the online short-gap identity-repair stage. Disabling this stage degrades the raw tracker from 82.25 to 80.98 IDF1, showing that it improves short-term continuity before offline repair. However, after applying the full offline pipeline, performance recovers to 84.94 IDF1, indicating that the final gain is still dominated by the offline relinking stage.

\subsection{Robustness to Threshold Selection}

A common concern with heuristic post-processing is that performance may depend on narrowly tuned threshold values. To examine this issue, we vary the principal thresholds of the stitching and relinking stages over broad operating ranges. Table~\ref{tab:sensitivity_summary} summarizes the resulting IDF1 variation. The tested ranges are chosen to span meaningfully looser and stricter settings than the default configuration rather than only local perturbations around a hand-tuned optimum.

\begin{table}[t]
\centering
\caption{Sensitivity summary for the main heuristic thresholds. Across most operating ranges, IDF1 varies only modestly, indicating that the repair backend is not critically dependent on narrow hand tuning.}
\label{tab:sensitivity_summary}
\small
\resizebox{\columnwidth}{!}{
\begin{tabular}{lccc}
\toprule
Parameter & Tested range & IDF1 range & Variation \\
\midrule
Stitch max gap & 10--60 & 84.93--84.93 & 0.00 \\
Stitch max distance & 40--120 & 84.63--84.93 & 0.30 \\
Stitch max angle & 30--60 & 84.93--84.93 & 0.00 \\
Stitch max speed ratio & 2.0--4.0 & 84.93--84.93 & 0.00 \\
Stitch min speed & 0.1--0.5 & 84.87--84.93 & 0.07 \\
Stitch velocity window & 2--20 & 84.03--84.93 & 0.91 \\
\midrule
Relink score threshold & 0.90--0.975 & 84.66--84.93 & 0.27 \\
Relink max gap & 30--90 & 84.92--84.93 & 0.01 \\
Relink max distance & 80--160 & 84.93--85.05 & 0.12 \\
Relink border margin & 10--40 & 84.46--84.93 & 0.47 \\
\bottomrule
\end{tabular}
}
\end{table}

Across most settings, IDF1 changes only marginally, typically by less than 0.5 percentage points. Noticeable degradation appears only for clearly over-conservative or over-smoothed configurations, such as a very large stitching velocity window or a large relinking border margin. These results indicate that the proposed repair backend is not brittle and that its gains do not depend on precise benchmark-specific tuning. In other words, the improvement is not tied to a narrow set of ``magic numbers,'' but persists across broad operating ranges for most thresholds.

\subsection{Qualitative Results}

Figure~\ref{fig:qualitative} visualizes representative tracking results across several challenging sequences. Qualitative examples suggest that the identity-repair backend produces smoother trajectories and fewer spurious identity breaks when pedestrians undergo temporary occlusion or cross paths. In particular, the repaired trajectories exhibit fewer mid-sequence identity breaks in cases where pedestrians undergo brief occlusion or motion-induced fragmentation.

\section{Discussion}

The ablation results indicate that, in thermal pedestrian MOT, identity continuity depends more on conservative trajectory relinking than on increasing online tracker complexity. This behavior is consistent with the low-information nature of thermal imagery: appearance cues are weak, while short-term motion and spatial continuity remain comparatively reliable. The strong effect of removing the spatial cue (Table~\ref{tab:component_ablation}, $-1.76$ pp) deserves particular attention. Unlike motion or border cues, spatial proximity operates as a hard geometric constraint that rules out implausible associations before any scoring takes place. In thermal scenes where pedestrian silhouettes are nearly interchangeable, this coarse spatial filter provides the most reliable evidence for identity, whereas motion direction and speed though informative are noisier due to stop-and-go pedestrian behavior. Accordingly, the contribution of this work is best understood not as a new tracking architecture, but as a controlled analysis revealing that identity continuity in thermal video is an emergent property of global scene-level spatial-temporal consistency rather than of local frame-to-frame association or individual detection quality.

At the same time, the study has limitations. All controlled ablations are conducted on a single thermal benchmark split, so the observed ranking of cues may not transfer unchanged to other sensing setups, frame rates, or crowd densities. In addition, while we cite existing lightweight offline association methods such as Translink~\cite{zhang2023translink} and AFLink~\cite{du2023strongsort} in our review of related work, we do not directly compare the proposed heuristic relinking backend against these methods on the same thermal dataset. Such a comparison would help clarify whether the observed gains are specific to our hand-crafted relinking design or whether similar improvements can be obtained with learned tracklet association under the same thermal conditions. Future work should include this comparison and further evaluate the same repair logic on additional thermal datasets to examine whether lightweight learned association can complement conservative relinking without sacrificing robustness. Overall, our findings suggest that identity preservation in thermal video is less dependent on appearance modeling and more closely tied to enforcing global scene coherence over time. We note that the comparison with other trackers is contextual, as the proposed post-processing is not applied to them.

\section{Conclusion}
This paper studies identity continuity in thermal video through a controlled analysis of trajectory-level recovery. Controlled ablations show that most of the identity-consistency improvement comes from conservative offline relinking, while many auxiliary thresholds remain stable across broad operating ranges. These results suggest that, in low-information thermal imagery, precise spatial-temporal relinking improves trajectory continuity more reliably than increasing online tracker complexity. We hope the presented baseline, ablation protocol, and empirical findings provide a useful reference for future thermal MOT research and for the design of lightweight identity-repair systems in low-visibility or flawed settings~\cite{lin2025clip,lin2025lfquiad, ahmar2024enhancing} .

\section*{Acknowledgments}
The authors gratefully acknowledge Celine Hsu for her assistance in preparing the figures and poster for this work. Her efforts significantly enhanced the clarity and visual presentation of the paper.

{
    \small
    \bibliographystyle{ieeenat_fullname}
    \bibliography{main}
}


\end{document}